\documentclass[]{article}
\usepackage[utf8]{inputenc}
\usepackage[letterpaper]{geometry}
\usepackage{amta2020}
\usepackage{times}
\usepackage{url}
\usepackage{latexsym}
\usepackage{natbib}
\usepackage{layout}

\usepackage{booktabs}
\usepackage{amsmath}
\usepackage{caption}
\usepackage{float}
\usepackage{pgfplots}
\usepackage{pgfplotstable}
\usepackage{tikz}
\usepackage{todonotes}
\usepackage{enumitem}
\usepackage{multirow}
\usepackage{siunitx}
\usepackage{subcaption}

\begin{document}

\title{\bf Domain Robustness in Neural Machine Translation}

\author{\name{\bf Mathias Müller} \hfill  \addr{mmueller@cl.uzh.ch}\\
        \name{\bf Annette Rios} \hfill \addr{rios@cl.uzh.ch}\\
        \name{\bf Rico Sennrich} \hfill \addr{sennrich@cl.uzh.ch}\\
        \addr{Department of Computational Linguistics, University of Zurich, Switzerland}
}

\maketitle
\pagestyle{empty}

\begin{abstract}
Translating text that diverges from the training domain is a key challenge for machine translation. Domain robustness---the generalization of models to unseen test domains---is low for both statistical (SMT) and neural machine translation (NMT). In this paper, we study the performance of SMT and NMT models on out-of-domain test sets. We find that in unknown domains, SMT and NMT suffer from very different problems: SMT systems are mostly adequate but not fluent, while NMT systems are mostly fluent, but not adequate. For NMT, we identify such hallucinations (translations that are fluent but unrelated to the source) as a key reason for low domain robustness. To mitigate this problem, we empirically compare methods that are reported to improve adequacy or in-domain robustness in terms of their effectiveness at improving domain robustness.
In experiments on German$\to$English OPUS data, and German$\to$Romansh (a low-resource setting) we find that several methods improve domain robustness. While those methods do lead to higher BLEU scores overall, they only slightly increase the adequacy of translations compared to SMT.

\end{abstract}

\section{Introduction}\label{sec:intro}

Even though neural models have improved the state-of-the-art in machine translation considerably in recent years, they still underperform in specific conditions.
One such condition is out-of-domain translation.
\citet{Koehn2017} found that neural machine translation (NMT) systems perform poorly in such settings and that their poor performance cannot be explained solely by the fact that out-of-domain translation is difficult: non-neural, statistical machine translation (SMT) systems were superior at this task.
For this reason, \citet{Koehn2017} identified translation of out-of-domain text as a key challenge for NMT.

Catastrophic failure to translate out-of-domain text can be viewed as overfitting to the training domain, i.e.\ systems learn idiosyncrasies of the domain rather than more general features.
Our goal is to learn models that generalize well to unseen data distributions, including data from other domains.
We will refer to this property of showing good generalization to unseen domains as \textbf{domain robustness}.

We consider domain robustness a desirable property of NLP systems, along with other types of robustness, such as robustness against adversarial examples (\citealt{Goodfellow2015}) or typos in the input \citep{Belinkov2018}.
While domain adaptation with small amounts of parallel or monolingual in-domain data has proven very effective for NMT \citep[e.g.][]{luong2015,sennrich-haddow-birch:2016:P16-11,R17-1049,li-EtAl:2019:WMT1}, the target domain(s) may be unknown when a system is built, and there are language pairs for which training data is only available for limited domains.
Hence, domain robustness of systems without any domain adaptation is not only of theoretical interest, but also relevant in practice. 

\begin{figure}

\begin{itemize}[leftmargin=30pt]
\item[\textbf{SRC}]  \textit{Aber geh subtil dabei vor.}
\item[\textbf{REF}] \textit{But be subtle about it.}
\item[\textbf{HYP}] \textit{Pharmacokinetic parameters are not significantly affected in patients with renal impairment (see section 5.2).}
\end{itemize}

\caption{Example illustrating how a German$\to$English NMT system trained on medical text \textit{hallucinates} the translation of an out-of-domain input sentence.}
\label{fig:hallucinated-example}
\end{figure}

Model architectures and training techniques have evolved since \citet{Koehn2017}'s study, and it is unclear to what extent this problem still persists. We therefore revisit the hypothesis that NMT systems exhibit low domain robustness. In preliminary experiments, we demonstrate that current models still fail at out-of-domain translation: BLEU scores drop drastically for test domains other than the training domain. However, the overall out-of-domain translation quality of NMT systems is now on par with SMT systems. 

An analysis of our baseline systems reveals that \textbf{hallucinated} content occurs frequently in out-of-domain translations (see Figure \ref{fig:hallucinated-example} for an example).
Several authors present anecdotal evidence for NMT systems occasionally falling into a \textit{hallucination} mode where translations are grammatically correct but unrelated to the source sentence \citep{Arthur2016,Koehn2017,Nguyen2018}.
Our manual evaluation shows  that hallucination is more pronounced in out-of-domain translation. We therefore expect methods that alleviate the hallucination problem to indirectly improve domain robustness.

As a means to reduce hallucination, we experiment with several techniques and assess their effectiveness in improving domain robustness: reconstruction \citep{Tu2017,niu-etal-2019-bi},  subword regularization \citep{Kudo2018}, neural noisy channel models \citep{li2016mutual,yee2019simple}, and defensive distillation \citep{papernot2016distillation}, as well as combinations of these techniques. The main contributions of this paper are:

\begin{itemize}
    \item we perform an analysis of SMT and NMT systems that confirms that while in-domain BLEU increased, domain robustness remains a major problem even with state-of-the-art Transformer architectures. Our comparison of SMT and NMT shows differences in how performance degrades in unseen domains: SMT mostly suffers in terms of fluency, while NMT tends to produce more fluent, but less adequate translations (hallucinations).
    \item we test several techniques related to adequacy, robustness, or out-of-domain translation in regard to their effectiveness in improving domain robustness in NMT. We find that several techniques are moderately successful, most notably reconstruction, which reduces the average percentage of hallucinations in out-of-domain test sets from 35\% to 29\%.
    \item we show that despite moderate improvements, domain robustness remains a challenge in NMT, and provide code and data sets to serve as baselines for future work.
    
\end{itemize}

\section{Data Sets}

We report experiments on two different translation directions: German$\to$English (DE$\to$EN) and German$\to$Romansh (DE$\to$RM).

\begin{table}
        \centering
        \small
        \begin{tabular}{llrlllr}
        \toprule
        \multicolumn{3}{c}{\textbf{DE--EN}} & & \multicolumn{3}{c}{\textbf{DE--RM}} \\
        \cmidrule{1-3} \cmidrule{5-7}
         domains & corpora & size & & domains & corpora & size \\ \cmidrule{1-3} \cmidrule{5-7}
          medical    & EMEA &  1.1m & & law & Allegra, & \\
          IT  & GNOME, KDE, PHP, Ubuntu, OpenOffice & 380k & & & Press Releases & 100k \\
          koran  & Tanzil &   540k & & blogs & Convivenza & 20k \\
          law  & JRC-Acquis &  720k \\
          subtitles  & OpenSubtitles2018 &  22.5m \\
          \bottomrule
        \end{tabular}
        \caption{Data sets common to all of our experiments. Size indicates number of sentence pairs.}
        \label{tab:corpora}
\end{table}

\subsection{German$\to$English}

For all DE$\to$EN experiments, we use the same corpora as \citet{Koehn2017}, available from OPUS \citep{Lison2016}.

We use corpora from OPUS to define five domains: \textit{medical}, \textit{IT}, \textit{koran}, \textit{law} and \textit{subtitles}. See Table \ref{tab:corpora} for an overview of sizes per domain. The domains are quite distant, and we therefore expect that systems trained on a single domain will have low domain robustness if tested on other domains.

For each domain, we select 2000 consecutive sentence pairs each for development and testing. Our test sets are different from \citet{Koehn2017}, so results are not directly comparable.
In all experiments, the \textit{medical} domain serves as the training domain, while the remaining four domains are used for testing.

\subsection{German$\to$Romansh}

To complement our DE$\to$EN experiments, we also train systems for DE$\to$RM. Romansh is a Romance language that, with an estimated \num{40000} native speakers, is low-resource, but has some parallel resources thanks to its status as an official Swiss language.
Domain robustness is of particular relevance in low-resource settings since training data is typically only available for few domains.
Our training data consists of \num{100000} sentence pairs, specifically the Allegra corpus by \citet{scherrer-cartoni-2012-trilingual} which contains mostly law text, and an in-house collection of government press releases. As test domain (unseen during training), we use blog posts from Convivenza\footnote{\url{https://www.suedostschweiz.ch/blogs/convivenza}}.
From both data sets we randomly select 2000 consecutive sentence pairs as test sets. 

\section{State-of-the-art Models Exhibit Low Domain Robustness}

In this section, we establish that current NMT systems exhibit low domain robustness by analyzing our baseline systems automatically and manually.

\subsection{Experimental Setup for Baseline Models} \label{subsec:baseline-setup}

We use Moses scripts for punctuation normalization and tokenization. We apply truecasing trained on in-domain training data. Similarly, we apply BPE \citep{Sennrich2016} with 32k (DE$\to$EN) and 16k (DE$\to$RM) merge operations learned from in-domain data. We train two baselines: \\

\noindent \textbf{NMT Baseline} A standard Transformer base model trained with Sockeye \citep{DBLP:journals/corr/VaswaniSPUJGKP17,Hieber2018}.\\

\noindent \textbf{SMT Baseline} A standard, phrase-based statistical model trained with Moses \citep{koehnmoses}, using mtrain \citep{laubli2018mtrain} as frontend with standard settings. \\

We always test on several domains, including the training domain. We use a beam size of 10 to translate test data. We report case-sensitive BLEU \citep{Papineni2002} scores on detokenized text,
computed with SacreBLEU \citep{post:2018:WMT}\footnote{SacreBLEU version signature: \texttt{BLEU+c.mixed+\#.1+s.exp+tok.13a+v.1.4.1}.}.

\subsection{Analysis of Baseline Systems} \label{subsec:baseline-analysis}

Tables \ref{tab:baselines-de-en} and \ref{tab:baselines-de-rm} show automatic evaluation results for all our baseline models. Neural models achieve good performance on the respective in-domain test sets (61.5 BLEU on \textit{medical} for DE$\to$EN; 52.5 BLEU on \textit{law} for DE$\to$RM), but on out-of-domain text, translation quality is clearly diminished, with an average BLEU of roughly 12 (DE$\to$EN) and 19 (DE$\to$RM). The following analysis will focus on our DE$\to$EN baseline systems.

Compared to results reported by \citet{Koehn2017}, NMT has improved markedly since their study was conducted, and is now on par with SMT in out-of-domain settings ($11.8$ BLEU versus $11.7$). However, on the in-domain test set, our NMT baseline outperforms the SMT baseline by 3 BLEU. This result suggests that higher in-domain performance does not guarantee better out-of-domain translations.

Unknown words constitute one possible reason for failing to translate out-of-domain texts. As shown in Table \ref{tab:oov}, the percentage of words that are not seen during training is much higher in all out-of-domain test sets. However, unknown words cannot be the only reason for low translation quality: The test sets with the lowest BLEU scores (\textit{koran} and \textit{subtitles}) actually have an out-of-vocabulary (OOV) rate similar to the \textit{IT} test set, where BLEU scores are much higher for both baseline models.

Additionally, our SMT baseline shows better generalization to some domains unseen at training time, while the average BLEU is comparable to the NMT baseline. In the \textit{IT} domain, the result is most extreme: the SMT system beats the neural system by 4.3 BLEU. This demonstrates that the low domain robustness of NMT is not (only) a data problem, but also due to the model's inductive biases.

\begin{table}
    \begin{subtable}[t]{0.45\textwidth}
        \centering
        \begin{tabular}[t]{lrr}
        \toprule
         & SMT  & NMT  \\  \cmidrule{2-3} 
          {\bfseries in-domain} \\
           medical   & 58.4 & \textbf{61.5} \\ \addlinespace
           {\bfseries out-of-domain} \\
           IT & \textbf{21.4} & 17.1 \\
           koran  & \textbf{1.4} & 1.1 \\
           law  & 19.8 & \textbf{25.3} \\
           subtitles  & \textbf{4.7} & 3.4 \\
           \midrule
           average (out-of-domain) & \textbf{11.8} & 11.7 \\
           \bottomrule
        \end{tabular}
        \caption{}
        \label{tab:baselines-de-en}
    \end{subtable}
    \hfill
    \begin{subtable}[t]{0.45\textwidth}
        \centering
        \begin{tabular}[t]{lrr}
        \toprule
         & SMT  & NMT  \\  \cmidrule{2-3} 
          {\bfseries in-domain} \\
           law   & 45.2 & \textbf{52.5} \\ \addlinespace
           {\bfseries out-of-domain} \\
           blogs & 15.5 & \textbf{18.9} \\
           \bottomrule
        \end{tabular}
        \caption{}
        \label{tab:baselines-de-rm}
    \end{subtable}
    \caption{BLEU scores of (a) baseline DE$\to$EN systems trained on {\em medical} data, (b) baseline DE$\to$RM systems trained on {\em law} data.}
\end{table}

\begin{table}
    \begin{subtable}[t]{0.40\textwidth}
        \centering
        \begin{tabular}[t]{lr}
        \toprule
         & OOV rate \\ \cmidrule{2-2}
            {\bfseries in-domain} \\
           medical   &  2.42\%\\ \addlinespace
           {\bfseries out-of-domain} \\
           IT & 20.09\%\\
           koran  & 18.63\%\\
           law  & 9.39\%\\
           subtitles  & 18.16\%\\
           \bottomrule
        \end{tabular}
        \caption{}
        \label{tab:oov}
    \end{subtable}
    \hfill
    \begin{subtable}[t]{0.50\textwidth}
        \centering
        \begin{tabular}[t]{lrr}
        \toprule
           & SMT & NMT \\ \cmidrule{2-3}
           {\bfseries in-domain} \\
           medical & 53.2 & \textbf{61.4} \\ \addlinespace
           {\bfseries out-of-domain} \\
           IT & 31.9 & \textbf{44.7} \\
           koran & 15.2 & \textbf{15.9} \\
           law  & 58.3 & \textbf{62.3} \\
           subtitles & 14.9 & \textbf{18.5} \\
           \midrule
           average (out-of-domain) & 30.1 & \textbf{35.4} \\
           \bottomrule
        \end{tabular}
        \caption{}
        \label{tab:baseline-general}
    \end{subtable}
    \caption{For the language pair DE$\to$EN: (a) Out-of-vocabulary (OOV) rates of in-domain and out-of-domain test sets. (b) BLEU scores of baselines trained on a concatenation of \textit{all} domains.}
\end{table}



As a further control, we train additional baseline systems trained on \textit{all} domains.\footnote{The \textit{subtitles} domain (23m sentences) was subsampled to 1m sentence pairs so as not to overwhelm the remaining domains (3m sentences in total). We also removed any overlap between the training and test sets.} We use it to test whether the data we have held out for out-of-domain testing is inherently more difficult to translate than the in-domain test set. The results in Table \ref{tab:baseline-general} show that this is not the case. For the NMT baseline, BLEU ranges between 15.9 and 62.3, with an average out-of-domain BLEU of 35.4.


\subsubsection{Hallucination} \label{subsubsec:hallucination}

NMT models can be understood as language models of the target language, conditioned on a representation of a source text.
This means that NMT models have no explicit mechanism---as SMT models do---that enforces coverage of the source sentence, and if the representation of an out-of-domain source sentence is outside the training distribution, it can be seemingly ignored.
This gives rise to a tendency to \textit{hallucinate} translations, i.e.\ to produce translations that are fluent, but unrelated to the content of the source sentence \citep{lee2019hallucinations}.

We hypothesize that hallucination is more common in out-of-domain settings. A small manual evaluation performed by the main author confirms that this is indeed the case. We evaluate the fluency and adequacy of our baselines (we refer to them as \textbf{NMT} and \textbf{SMT}). In a blind setup, we annotate a random sample of 100 sentence pairs per domain. As controls, we mix in pairs of \texttt{(source, actual reference)}, treating the reference translation as an additional system.

\textbf{Evaluation of adequacy} The annotator is presented with a sentence pair and asked to judge whether the translation is adequate, partially adequate or inadequate. 


\textbf{Evaluation of fluency} We use the same data as for the evaluation of adequacy, however, the annotator is shown only the translation, without the corresponding source sentence. The annotator is asked whether the given sentence is fluent, partially fluent or not fluent.

\begin{figure}
    \centering
    \begin{subfigure}[t]{0.45\textwidth}%
        \centering
        \includegraphics[height=3.7cm]{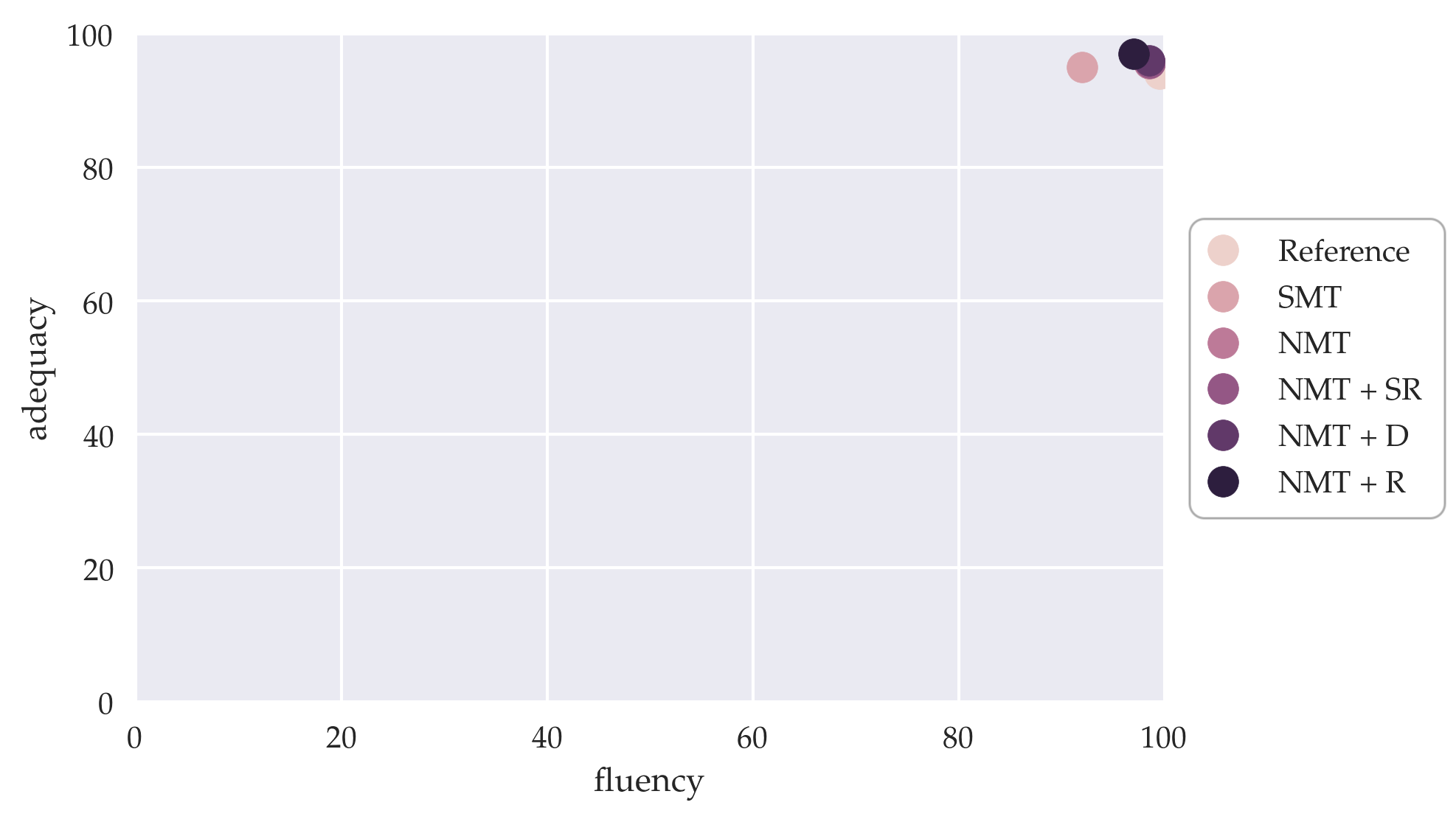}%
        \caption{in-domain}
        \label{fig:manual_eval_baseline}
    \end{subfigure}%
    \hspace*{\fill}%
    \begin{subfigure}[t]{0.45\textwidth}%
        \centering
        \includegraphics[height=3.7cm]{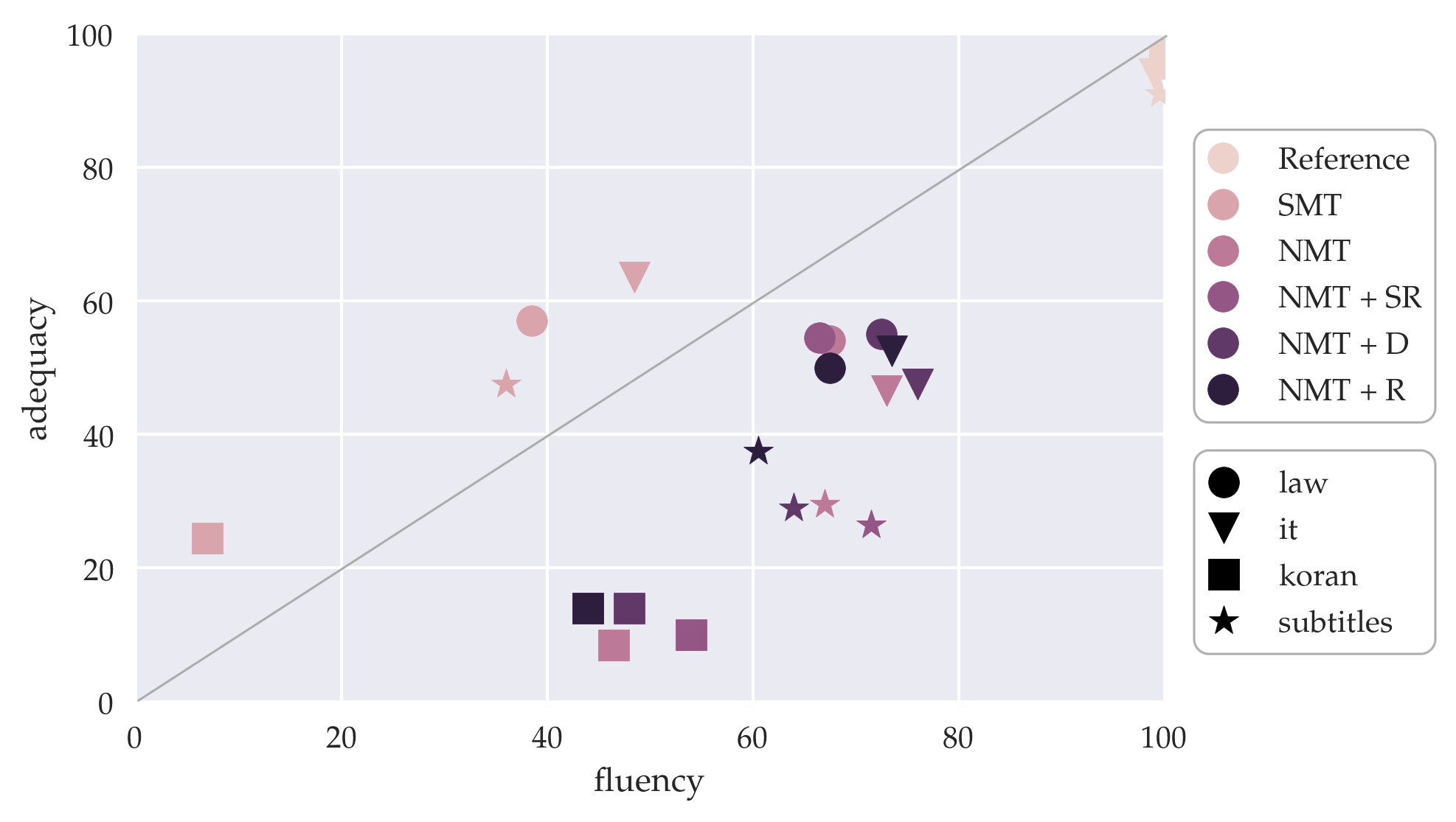}%
        \caption{out-of-domain}
        \label{fig:manual_eval_more}
    \end{subfigure}%
    \caption{Manual evaluation of adequacy and fluency for DE$\to$EN. Legend: marker colors are different systems, marker types are different domains. SR=Subword Regularization, D=Distillation, R=Reconstruction}
    \label{fig:manual_eval}
\end{figure}

Figure \ref{fig:manual_eval} shows the results of the manual evaluation in terms of adequacy and fluency. For visualization, individual fluency values are computed as follows:

\begin{equation*}
    1.0 * n_f + 0.5 * n_p + 0.0 * n_n
\end{equation*}

Where $n_f$, $n_p$ and $n_n$ are the number of fluent, partially fluent and non-fluent translations, respectively. Adequacy values are computed in the same way. On the in-domain test set, both baselines achieve high adequacy and fluency, with the NMT baseline effectively matching the adequacy and fluency of the reference translations.

Regarding adequacy, the in-domain samples contain only a small number of translations with content unrelated to the source (1\% to 2\%). On out-of-domain data, on the other hand, both baselines produce a high number of inadequate translations: 57\% (SMT) and 84\% (NMT). These results suggest that the extremely low BLEU scores on these two test sets (see Table \ref{tab:baselines-de-en}) are in large part due to made up content in the translations. 

Regarding fluency in out-of-domain settings, SMT and NMT baselines behave very differently: SMT translations are more adequate, while NMT translations are more fluent. This trend is most extreme in the \textit{koran} domain, where only 2\% of SMT translations are found to be fluent (compared to 36\% for NMT).

Further analysis of both annotations shows that NMT translations found to be inadequate are not necessarily disfluent in out-of-domain settings. Table \ref{tab:ratios} shows that, on average, on out-of-domain data, 35\% of NMT translations are both inadequate and fluent, while the same is only true for 4\% of SMT translations. We refer to translations of this kind as \textbf{hallucinations}.

\begin{table}
    \centering
    \begin{tabular}{lrr}
    \toprule
    & in-domain & OOD average \\
    \cmidrule{2-3}
    Reference & 2\% & 2\% \\
    \midrule
    NMT & 2\% & 35\% \\
    SMT & 1\% & 4\% \\
    \midrule
    NMT + Subword Regularization & 1\% & 37\% \\
    NMT + Distillation & 3\% & 33\% \\
    NMT + Reconstruction & 1\% & 29\% \\
    \bottomrule
    \end{tabular}
    \caption{Percentage of translations judged to be hallucinations (both \textbf{not adequate} and at least \textbf{partially fluent}) in the manual evaluation.}
    \label{tab:ratios}
\end{table}

To summarize our analysis of baseline models, we find that the domain robustness of current NMT systems is still lacking and that inadequate, but fluent translations are a prominent issue. This motivates our choice of techniques to improve domain robustness.

\section{Approaches to Improve Domain Robustness} \label{sec:approaches}

We discuss approaches that can potentially remedy the problem of low domain robustness, and compare them in subsequent experiments.


\subsection{Subword Regularization} \label{subsec:subword-reg}

Subword regularization \citep{Kudo2018} is a form of data augmentation that, instead of applying a fixed subword segmentation like BPE, probabilistically samples a new subword segmentation for each epoch. At test time, the model either uses 1-best segmentation, or translates the k-best segmentations and selects the highest-probability translation.

\citet{Kudo2018} reports large improvements on low-resource and out-of-domain settings.
In particular, improvements on in-house patent, web, and query test sets were in the range of 2--10 BLEU.
In this work, we evaluate subword regularization on public datasets.
We apply sampling at training time and translate 1-best segmented sentences at test time.

\subsection{Defensive Distillation} \label{subsec:defensive-distillation}

We hypothesize that defensive distillation can be used to improve domain robustness. Defensive distillation exploits \textit{knowledge distillation} to fend off adversarial attacks.

Knowledge distillation is a technique to derive models from existing ones, instead of training from scratch. The idea was introduced for simple image classification models by \citet{ba2014deep} and \citet{hinton2015distilling}. A first model (called the teacher) is trained in the usual fashion. Then, a second model (called the student) is trained using the predictions of the teacher model instead of the labels in the training data.


Typically, knowledge distillation is used to approach the performance of a complex teacher model (or ensemble of models) with a simpler student model.
Another application is \textit{defensive distillation} (e.g. \citealt{papernot2016distillation}), where the student shares the network architecture with the teacher, with the purpose not being model compression, but improving the model's generalization to samples outside of its training set, and specifically robustness against adversarial examples \citep{szegedy2013intriguing,goodfellow6572explaining}.

Defensive distillation has been shown to be effective at improving robustness to adversarial examples in image recognition.
In this work, we apply it to NMT and test its effect on domain robustness, which \citet{papernot2016distillation} hint at but do not empirically test.
As proposed by \citet{kim-rush:2016:EMNLP2016}, we train the student model on teacher translations produced with beam search, instead of training on soft prediction labels.


\subsection{Reconstruction} \label{subsec:reconstruction}

Reconstruction \citep{Tu2017} is a change to the model architecture that addresses the problem of adequacy. The authors propose to extend encoder-decoder models with a \textit{reconstructor} component that learns to reconstruct the source sentence from decoder states.
The reconstructor has two uses: as a training objective, it forces the decoder representations to retain information that will be useful for reconstruction; during inference, it can provide scores that can be used for reranking.

However, we observed in initial experiments that reconstruction from hidden states can be too easy: the reconstruction loss on training batches diminishes very quickly, to the point of being insignificant. To prevent the model from simply reserving parts of the decoder hidden states to memorize the input sentence, we use reconstruction from actual translations instead of hidden states \citep{niu-etal-2019-bi}. Translations are produced with differentiable sampling via the Straight-Through Gumbel Softmax \citep{JanGuPoo17}, which still allows joint optimization of translation and reconstruction.
While \citet{niu-etal-2019-bi} implement reconstruction for recurrent architectures, we apply the technique to Transformers.

In order to avoid introducing any additional parameters for reconstruction, as recommended in \citet{niu-etal-2019-bi}, we train a multilingual, bi-directional system with shared parameters as a further baseline. This bi-directional system is used to initialize the fine-tuning of reconstruction models. We empirically test whether our bilingual baseline and this multilingual baseline have comparable performance.

\subsection{Neural Noisy Channel Reranking}

Even though the methods we presented previously do lead to improved out-of-domain translation quality, the models still suffer from low adequacy. Also, our reconstruction models only perform reconstruction during training, and the reverse translation direction is not exploited, for instance by reranking translations \citep{Tu2017}.


We conjecture that this problem can be addressed with a neural noisy channel model \citep{li2016mutual}. Standard NMT systems only model $p(y|x)$, which can lead to a ``failure mode that can occur in conditional models in which inputs are explained away by highly predictive output prefixes'' \citep{DBLP:conf/iclr/YuBDGK17}. Noisy channel models propose to also model $p(x|y)$ and $p(y)$ to alleviate this effect.

In practical terms, noisy channel models are implemented by modifying the core decoding algorithm, or simply as n-best list reranking. We adopt the latter,
since n-best list reranking was shown to have equal or better performance than more computationally costly methods that score partial hypotheses during beam search \citep{yee2019simple}.

\section{Experimental Setup for Proposed Methods}

This section describes how we preprocessed data and trained the models described in Section \ref{sec:approaches}.
Unless stated otherwise, the data is preprocessed in the same way as for the baselines (see Section \ref{subsec:baseline-setup}).

\textbf{Subword Regularization Models} We integrate subword regularization in Sockeye, using the Python library provided by \citet{Kudo2018}\footnote{\url{https://github.com/google/sentencepiece}}. The training data is \textbf{not} segmented with BPE in this case. Instead, the training tool is given truecased data, and new segmentations are sampled before each training epoch. We use the following hyperparameters: we set the smoothing parameter $\alpha$ to 0.1 and use an n-best size of 64. For the validation and test data we use 1-best segmentation.

\textbf{Defensive Distillation Models} 
We use our baseline Transformer model as the teacher model. We translate the original training set with beam size 10.
The student is trained on the translations of the teacher model using the same hyperparameters and being initialized with the parameters of the teacher model.

\textbf{Reconstruction Models}
We implement differentiable sampling reconstruction for Transformer models in Sockeye and release the implementation.\footnote{
\url{https://github.com/ZurichNLP/sockeye/tree/domain-robustness}
}
We first train a multilingual Transformer model using the approach of \citet{johnson-etal-2017-googles}.


After early stopping, we continue training with reconstruction as an additional loss.
All hyperparameters remain the same, except for the new loss and a lower initial learning rate. For testing we select the model with the lowest validation perplexity.
We use the reconstruction loss only for training, not for reranking.

\textbf{Noisy Channel Reranking}
For each hypothesis, we store an n-best list of 50. We produce the following scores: $p(y|x)$ (usual translation score), $p(x|y)$ (translation score in reverse direction) and $p(y)$ (language model score in target language). $p(y|x)$ and $p(x|y)$ are computed with the same model since it is bi-directional.

In order to produce $p(y)$ scores we train a Transformer language model with fairseq \citep{ott2019fairseq} using standard settings. We impose a large penalty of $-100$ for hypotheses that contain subwords not found in the target side training data.

The final hypothesis score for reranking is computed as a weighted multiplication:

\begin{equation*}
\text{score}(x,y) =  p(y|x)^{\lambda_{tf}} * p(x|y)^{\lambda_{tb}} * p(y)^{\lambda_{lm}}
\end{equation*}

The best weights are found with simple grid search on the in-domain development set, with ${\lambda_{tf}} \in [0.0,0.1,...,1.0]$, ${\lambda_{lm}} \in [0.0,0.01,...,0.1,0.2,...,1.0]$, and $\lambda_{tb}=1-\lambda_{tf}-\lambda_{lm}$.
The best weight combination is then used to compute scores and perform reranking for the test data of all domains. Table \ref{tab:grid-weights} lists optimal weights found for each model individually.


\begin{table}
    \centering
    \begin{tabular}{lrrrcrrr}
    \toprule
    & \multicolumn{3}{c}{\textbf{DE$\to$EN}} & &  \multicolumn{3}{c}{\textbf{DE$\to$RM}} \\
    \cmidrule{2-4} \cmidrule{6-8}
    & $\lambda_{tf}$ & $\lambda_{tb}$ & $\lambda_{lm}$ & & $\lambda_{tf}$ & $\lambda_{tb}$ & $\lambda_{lm}$ \\
    \cmidrule{2-4} \cmidrule{6-8}
    (5) Multilingual & 0.7 & 0.26 & 0.04 & & 0.9 & 0.09 & 0.01 \\
    (6) Reconstruction & 0.6 & 0.32 & 0.08 & & 0.9 & 0.09 & 0.01 \\
    \midrule
    (7) Multilingual + SR & 0.5 & 0.42 & 0.08 & & 0.9 & 0.09 & 0.01 \\
    (8) Reconstruction + SR & 0.5 & 0.46 & 0.04 & & 0.9 & 0.09 & 0.01 \\
    \bottomrule
    \end{tabular}
    \caption{Best weights for noisy channel reranking found with grid search on in-domain development set. Row numbers correspond to the ones in Table \ref{tab:results}. $\lambda_{tf}$=forward translation weight, $\lambda_{tb}$=backward translation weight, $\lambda_{lm}$= language model weight}
    \label{tab:grid-weights}
\end{table}

\section{Results}

We evaluate models in terms of BLEU on different domains (see Section \ref{subsec:eval-automatic}) and also annotate a subset of translations manually for fluency and adequacy (see Section \ref{subsec:eval-manual}).

\subsection{Automatic evaluation}
\label{subsec:eval-automatic}

Table \ref{tab:results} shows the results of our automatic evaluation. Overall, the proposed methods are able to outperform the SMT and NMT baselines but not in a consistent manner across domains or data conditions: an increase in in-domain BLEU does not always mean a similar increase on out-of-domain data. Similarly, an increase in BLEU in high-resource conditions (DE$\to$EN) does not consistently lead to better results in low-resource conditions (DE$\to$RM). 

\begin{table*}
    \small
    \centering
    \begin{tabular}{lrrcrr}
    \toprule
    & \multicolumn{2}{c}{\textbf{DE$\to$EN}} & &  \multicolumn{2}{c}{\textbf{DE$\to$RM}} \\
    \cmidrule{2-3} \cmidrule{5-6}
    & in-domain & average OOD & & in-domain & average OOD \\
    \cmidrule{2-3} \cmidrule{5-6}
    (1) SMT & 58.4 & 11.8 & & 45.2 & 15.5 \\
    (2) NMT & 61.5 & 11.7 & & 52.5 & 18.9 \\
    \midrule
    (3) NMT + SR & 61.4 & 11.2 & & \textbf{53.7} & 20.1 \\
    (4) NMT + D & 61.1 & 13.1 & & 52.5 & 19.3 \\
    \midrule
    (5) Multilingual & 61.4 & 11.7 & & 52.8 & 19.6 \\
    (6) Reconstruction & 61.5 & 12.5 & & 53.4 & 21.2 \\
    \midrule
    (7) Multilingual + SR & 60.3 & 12.8 & & 52.4 & 20.1 \\
    (8) Reconstruction + SR & 60.3 & \textbf{13.2} & & 52.4 & 20.3 \\
    \midrule
    (9) Multilingual + NC & 62.7 & 11.8 & & 53.1 & 21.4 \\
    (10) Reconstruction + NC & \textbf{62.8} & 13.0 & & 53.3 & \textbf{21.6} \\
    \midrule
    (11) Multilingual + SR + NC & 60.7 & 12.3 & & 53.1 & 21.4 \\
    (12) Reconstruction + SR + NC & 60.8 & 13.1 & & 52.4 & 20.7 \\
    \bottomrule
    \end{tabular}
    \caption{BLEU scores (higher is better) of all systems on test data. SR=Subword Regularization, D=Distillation, NC=Noisy Channel Model, average OOD=average BLEU score over out-of-domain test sets.}
    \label{tab:results}
\end{table*}

\textbf{Subword regularization} The results for subword regularization are mixed (see Row 3 in Table \ref{tab:results}). For DE$\to$EN, in-domain translation quality is comparable to the NMT baseline, while the average out-of-domain BLEU falls short of the NMT baseline (-0.5 BLEU). However, in the low-resource condition (DE$\to$RM), subword regularization improves both in-domain and out-of-domain translation (+1.2 in both cases).

This result is surprising given the larger gains reported by \citet{Kudo2018}. To validate our implementation of subword regularization, we reproduce an experiment from \citet{Kudo2018} with English-Vietnamese data from IWSLT 15 (see Table \ref{tab:iwslt}). With subword regularization we observe an improvement of 0.8 BLEU, which is lower than the 2 BLEU improvement reported by \citet{Kudo2018}, but we also note that our baseline model is stronger.

\begin{table}
    \centering
    \begin{tabular}{lcc}
    \toprule
    & Baseline & Subword \\
    & (BPE) & Regularization \\
    \cmidrule{2-3}
    \citet{Kudo2018} & 25.6 &  27.7 (+ 2.1) \\
    Our results & 28.3 & 29.1 (+ 0.8) \\
        \bottomrule
    \end{tabular}
    \caption{Reproducing results from \citet{Kudo2018} on IWSLT 15 English-Vietnamese data.}
    \label{tab:iwslt}
\end{table}

If subword regularization is combined with multilingual or reconstruction models (see Rows 7 and 8 in Table \ref{tab:results}), we observe no improvements on in-domain test sets, but gains on 3 out of 4 out-of-domain data sets, indicating that subword regularization is in fact helpful for domain robustness.

\textbf{Defensive Distillation} Distilling the training data also leads to improvements in BLEU on out-of-domain text (see Row 4 in Table \ref{tab:results}). The average gain is +1.4 for DE$\to$EN, but only +0.4 for DE$\to$RM. In-domain performance is either comparable or slightly worse (-0.4 BLEU for DE$\to$EN) than the NMT baseline. The technique was originally shown to guard against adversarial attacks, where inputs are only infinitesimally different from training examples. Our results indicate that generalization to out-of-domain inputs -- that are farther from the training data -- is similarly improved.

\textbf{Reconstruction} Since reconstruction models are fine-tuned from multilingual models, we report scores for those multilingual models as well. Row 5 of Table \ref{tab:results} shows that our multilingual models perform equally well or better than the NMT baseline. As shown in Row 6 of Table \ref{tab:results}, reconstruction outperforms the NMT baseline on out-of-domain data for both language pairs (+0.8 BLEU for DE$\to$EN, +2.3 BLEU for DE$\to$RM), while maintaining (DE$\to$EN) or improving (+0.9 BLEU for DE$\to$RM) in-domain BLEU.

\textbf{Noisy Channel Reranking} We evaluate the performance of noisy channel reranking in four different settings: applied to multilingual or reconstruction systems, both with and without subword regularization. The results are shown in Rows 9 to 12 of Table \ref{tab:results}.

For DE$\to$EN, reranking a reconstruction model achieves a good in-domain BLEU (+1.3 over the baseline), and slightly improves out-of-domain translation on average (+0.5 BLEU over reconstruction). For DE$\to$RM in our low-resource setting, reranking with a noisy channel model improves the reconstruction model by +0.4 BLEU, producing the best result overall. The improvement on out-of-domain translation is much larger for the multilingual model (+1.8 over multilingual model without reranking). Combining reranked models (see Rows 11 and 12 in Table \ref{tab:results}) with subword regularization does not lead to consistent improvements. Out-of-domain BLEU for DE$\to$RM is slightly better compared to a subword regularization system without reranking (+0.4 BLEU), while all other scores are comparable or worse.

We believe that the success of reranking could be limited for two reasons. Firstly, model weights $\lambda$ are optimized on in-domain data. Oracle experiments that optimize weights on out-of-domain data show that optimal model weighting differs greatly between domains and could further improve out-of-domain results by 0.5--1 BLEU on average. Secondly, reranking could be held back by the lack of diversity among hypotheses in n-best lists.


\subsection{Manual evaluation}
\label{subsec:eval-manual}

In a small manual evaluation, we analyze if methods that improve domain robustness in terms of BLEU also directly address the lack of adequacy in out-of-domain translation (see Section \ref{subsubsec:hallucination}). The results are shown in Figure \ref{fig:manual_eval_more}. For anectodal examples of translations by several models, see Table \ref{tab:hall-examples}.

\begin{table}
    \centering
    \small
    \begin{tabular}{lcl}
    \toprule
    Source & & - die Produktion in der Türkei entspricht 1,3 \% der chinesischen Produktion;\\
    Target & & - Turkey’s volume of production amounts to 1,3 \% of Chinese production, \\
    \midrule
    NMT Baseline & & - the production in slkei is 1.3\% of a Chinese \textbf{hamster ovary (CHO) cell}
\\
    Multilingual & & - production in turkei is equivalent to 1.3\% of Chinese \textbf{Hamster} production; \\
    Reconstruction & & - the production in thekei is equivalent to 1.3\% of the Chinese production; \\
    Reconstruction + NC & & - production in the turkei equals 1.3\% of the Chinese production;\\
    \bottomrule
    \end{tabular}
    \caption{Example translations for DE$\to$EN. Hallucinated parts are shown in \textbf{bold}.}
    \label{tab:hall-examples}
\end{table}

\textbf{Techniques with a bias for fluency} We find that subword regularization increases fluency by several percentage points in some domains. However, it fails to improve adequacy which explains why it consequently fails to improve out-of-domain BLEU. Defensive distillation improves translations in a similar way, with a bias for fluency while not consistenly improving the adequacy of out-of-domain translations.

\textbf{Reconstruction having a bias for adequacy} We show that reconstruction is limiting hallucination on out-of-domain data, reducing the percentage of inadequate translations by 5 percentage points on average. We also note that there appears to be a tradeoff between adequacy and fluency: while reconstruction does improve out-of-domain adequacy, the improvement comes at the cost of lower fluency.

\section{Discussion}


Overall, we emphasize that SMT no longer appears to have higher domain robustness than NMT---a widely held belief given the findings of \cite{Koehn2017}. However, SMT and NMT models achieve comparable domain robustness in different ways: SMT translations are more adequate (since SMT decoding implicitly enforces coverage), while NMT translations are more fluent (since an NMT decoder is simply a language model of the target language conditioned on source context). Therefore, evaluating domain robustness with automatic metrics only can be misleading and hide the fact that models have very different inductive biases, even if their BLEU score is similar.


Our experiments to reduce NMT hallucination and improve adequacy can be summarized as follows. Several methods lead to moderate improvements in translation quality, but only reconstruction combined with noisy channel reranking robustly increased BLEU across domains and data conditions (see Row 10 of Table \ref{tab:results}). Other techniques are successful only in certain conditions. For instance, combining multilingual systems with Subword Regularization improves out-of-domain translation only, but not in-domain quality (see Rows 7--8 of Table \ref{tab:results}). This suggests that techniques found to work well on in-domain test data cannot be assumed to have a similar effect on out-of-domain data without proper testing.


Our manual evaluation suggests that for NMT systems, an increase in out-of-domain BLEU also means that translations are slightly more adequate. However, NMT models still have a strong bias for fluency (all models under the diagonal in Figure \ref{fig:manual_eval_more}) and their adequacy falls short of the adequacy of SMT systems. Our most successful reconstruction model does indeed improve adequacy, but at the same time decreases fluency to a certain extent. We believe radically different approaches are needed to increase the coverage and adequacy of NMT translations without sacrificing their fluency.

\section{Conclusions}


Current NMT systems exhibit low domain robustness, i.e.\ they underperform if they are tested on a domain that differs strongly from the training domain.
This is especially problematic in settings where explicit domain adaptation is impossible because the target domain is unknown, or because we are in a low-resource setting where training data is only available for limited domains. We find that hallucinated translations are a common problem for NMT models in out-of-domain settings, which partially explains their low domain robustness.
Our results show that several methods yield improved generalization to out-of-domain data. We achieve an improvement in average out-of-domain BLEU of 1.5 (DE$\to$EN) and 2.7 (DE$\to$RM), as well as a reduction in hallucinated translations according to manual evaluation.


We analyzed the fluency and adequacy of translations manually, leading to a discussion of several pitfalls regarding the evaluation of NMT models in out-of-domain settings. Also, we believe that future research will need to address the lack of adequacy without losing fluency.
We consider domain robustness an unsolved problem and encourage further research. For this purpose, we share data and code to serve as a baseline for future experiments.\footnote{
\url{https://github.com/ZurichNLP/domain-robustness}
}







%

\small

\section*{Acknowledgements}

We are grateful to Manuela Weibel, Beni Ruef, Florentin Lutz and the \textit{Lia Rumantscha} for providing Rumansh data, to Xing Niu for his help with reconstruction training, and to \textit{Fremde Welten} for their help with manual annotations. MM and AR have received funding from the  Swiss  National  Science Foundation (SNF, grant number 105212\_169888). RS has received funding from the European Union’s Horizon 2020 Research and Innovation Programme (ELITR, grant agreement 825460).



\clearpage

\bibliographystyle{apalike}
\bibliography{domain}

\clearpage




\end{document}